%% file: main.tex
\documentclass{INTERSPEECH2023}

\interspeechcameraready 
\usepackage[acronym,toc,shortcuts,nonumberlist]{glossaries}
\usepackage{multirow}
\usepackage{subcaption}
\usepackage{color}
\usepackage{csquotes}  
\usepackage[table]{xcolor}
\usepackage{cleveref}
\input{acronyms}

\title{Mixture Encoder for Joint Speech Separation and Recognition}
\name{
  Simon Berger$^{1,2}$, Peter Vieting$^{1}$, Christoph Boeddeker$^{3}$, Ralf Schl\"uter$^{1,2}$, Reinhold Haeb-Umbach$^{3}$
}
\address{
  $^1$Machine Learning and Human Language Technology,\\Computer Science Department, RWTH Aachen University, Germany\\
  $^2$AppTek GmbH, Germany\\
  $^3$Paderborn University, Germany}
\email{\{sberger,vieting,schlueter\}@cs.rwth-aachen.de, \{boeddeker,haeb\}@nt.upb.de}

\begin{document}

\maketitle

\begin{abstract}
	Multi-speaker \gls{ASR} is crucial for many real-world applications, but it requires dedicated modeling techniques. Existing approaches can be divided into modular and end-to-end methods. Modular approaches separate speakers and recognize each of them with a single-speaker \gls{ASR} system. End-to-end models process overlapped speech directly in a single, powerful neural network. This work proposes a middle-ground approach that leverages explicit speech separation similarly to the modular approach but also incorporates mixture speech information directly into the \gls{ASR} module in order to mitigate the propagation of errors made by the speech separator. We also explore a way to exchange cross-speaker context information through a layer that combines information of the individual speakers. Our system is optimized through separate and joint training stages and achieves a relative improvement of 7\% in \acrlong{WER} over a purely modular setup on the SMS-WSJ task.
\end{abstract}

\glsresetall

\section{Introduction}
Multi-speaker \gls{ASR} describes the task of automatically transcribing speech in a scenario where multiple speakers speak at the same time and it is highly relevant for many applications.
From a classical cocktail party scenario to meetings in everyday life, natural conversations typically contain overlapped speech.
Transcribing this type of data requires dedicated techniques as standard single-speaker \gls{ASR} methods do not perform well in this context \cite{menne2019deepclustering}.

Existing works for multi-speaker \gls{ASR} can be divided into two main lines of research.
First, there are modular approaches \cite{menne2019deepclustering, neumann2020joint} which deploy a speech separation frontend and an \gls{ASR} backend.
The frontend separates the mixture signal into multiple signals that each contain only a single speaker.
These can subsequently be recognized by a standard single-speaker \gls{ASR} system.
On the other hand, there are end-to-end approaches \cite{seki2018purely, kanda2020sot, sklyar2021msrnnt} that process the overlapped speech directly using one large \gls{NN} without explicit separation.

Modular approaches have the advantage of a clearly interpretable division between the components.
This allows to listen to the separated speakers and enables an easier analysis of a system.
At the same time, the \gls{ASR} module relies on an imperfect separation produced by the separator.
Since hard decisions are already enforced on this intermediate level, errors are propagated and cannot easily be corrected by the \gls{AM}.

In contrast, end-to-end models avoid this by having only one global decision.
However, they require more powerful \glspl{NN} and do not utilize the strengths of existing speaker separation methods.
Furthermore, they have to rely on \gls{PIT} \cite{kolbaek2017uPITBLSTM} or use heuristic assignment policies for the correct mapping of label posteriors to transcriptions in \gls{ASR} training.
This can lead to high computational costs and is stated to be less reliable than solving the permutation problem on signal level \cite{neumann2020joint}.

In this work, we propose an approach that can be viewed as a middle course between these different methods.
We leverage explicit speech separation by incorporating a classical frontend that is pre-trained in a permutation invariant way.
In addition, the proposed mixture encoder processes the mixture speech signal and encodes information that is fed into the \gls{AM} in addition to the separated signals.
This information can be used to mitigate the error propagation effect.
Since we train on simulated (artificially overlapped) data, the label ambiguity problem can be avoided by solving the permutation using the output of the pre-trained speech separator and the reference single speaker signals during \gls{AM} training.
In addition to the separate training of separator and \gls{AM}, we optimize the whole system in a final joint training stage.
We show improvements in \gls{ASR} accuracy using this proposed approach over a purely modular setup.

\begin{figure*}[h]
	\begin{subfigure}[t]{.24\textwidth}
		\centering
		\includegraphics[width=\columnwidth]{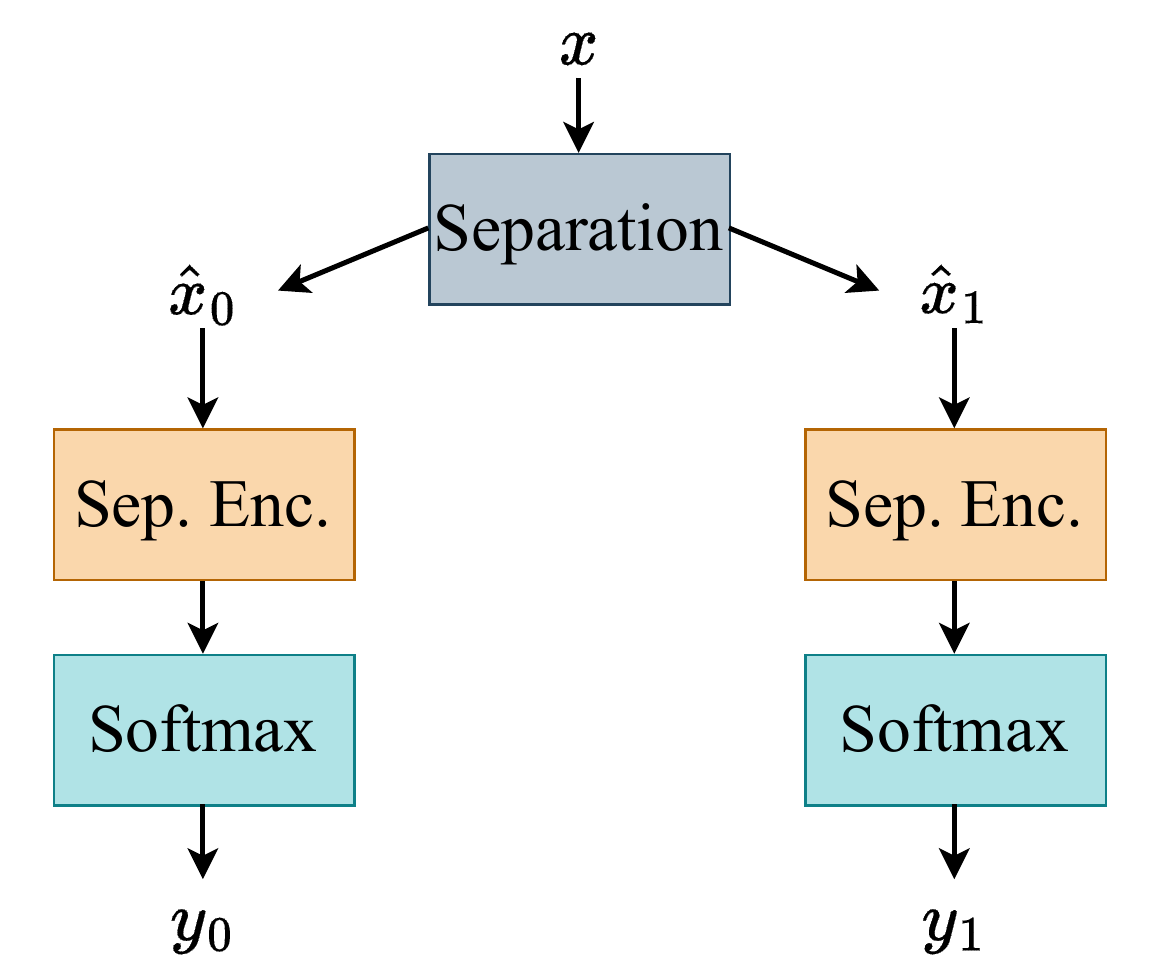}
		\caption{Modular structure.}
		\label{fig:modular}
	\end{subfigure}
	\hfill
	\begin{subfigure}[t]{.24\textwidth}
		\centering
		\includegraphics[width=\columnwidth]{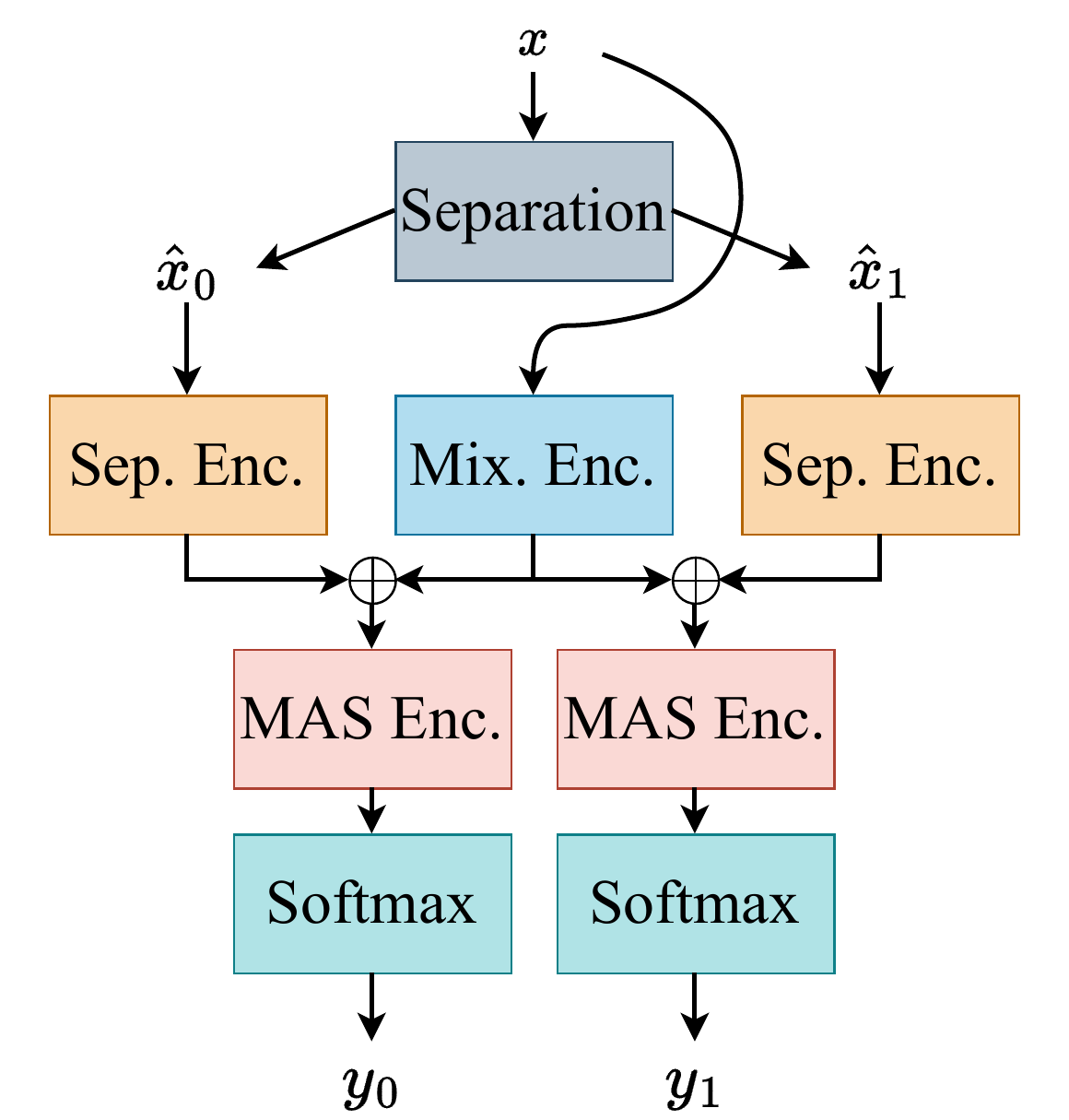}
		\caption{Structure with mixture encoder and \acrshort{MAS} encoders.}
		\label{fig:nocombine}
	\end{subfigure}
	\hfill
	\begin{subfigure}[t]{.24\textwidth}
		\centering
		\includegraphics[width=\columnwidth]{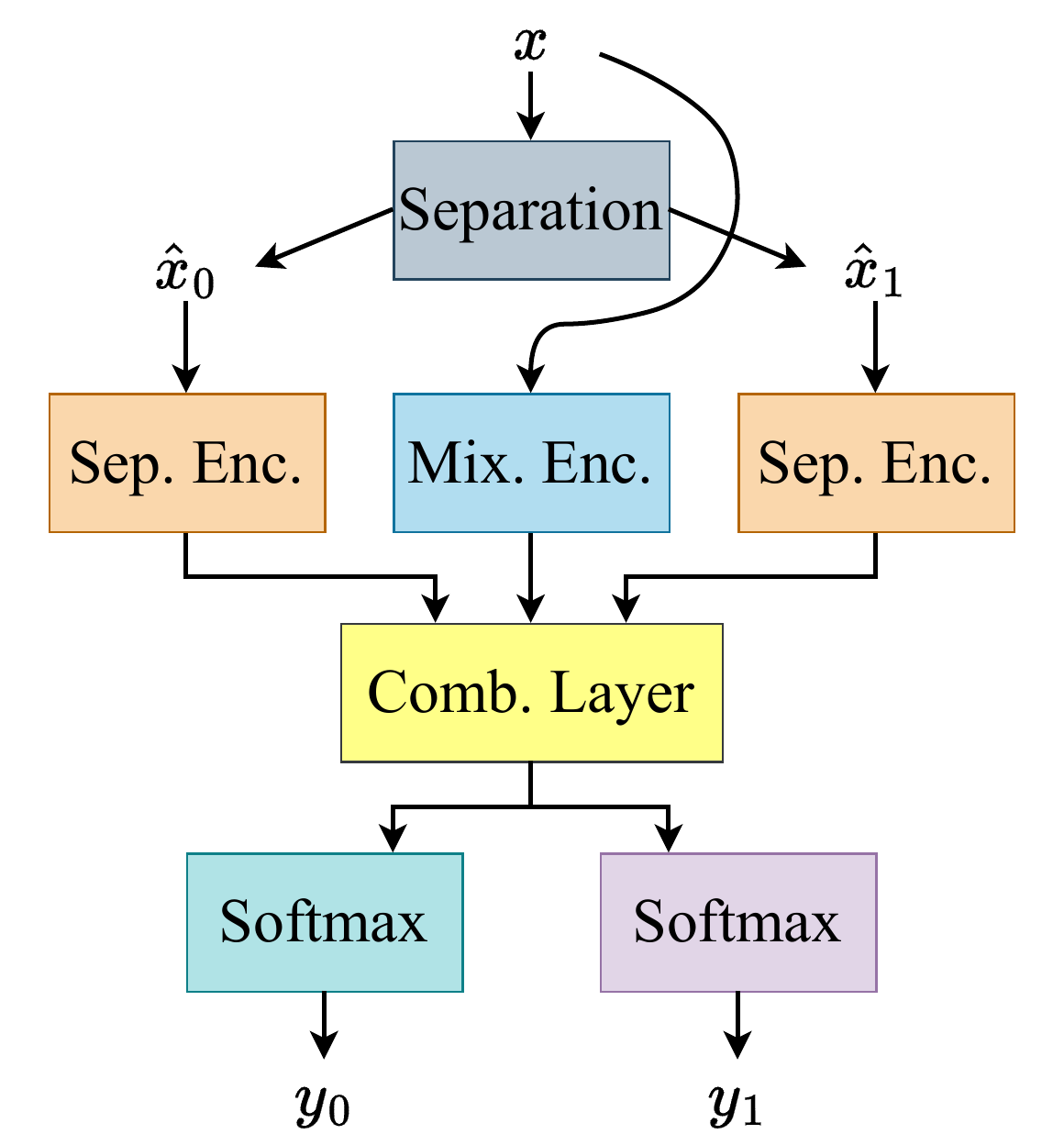}
		\caption{Structure with combine layer instead of \acrshort{MAS} encoders.}
		\label{fig:nomas_combine}
	\end{subfigure}
	\hfill
	\begin{subfigure}[t]{.24\textwidth}
		\centering
		\includegraphics[width=\columnwidth]{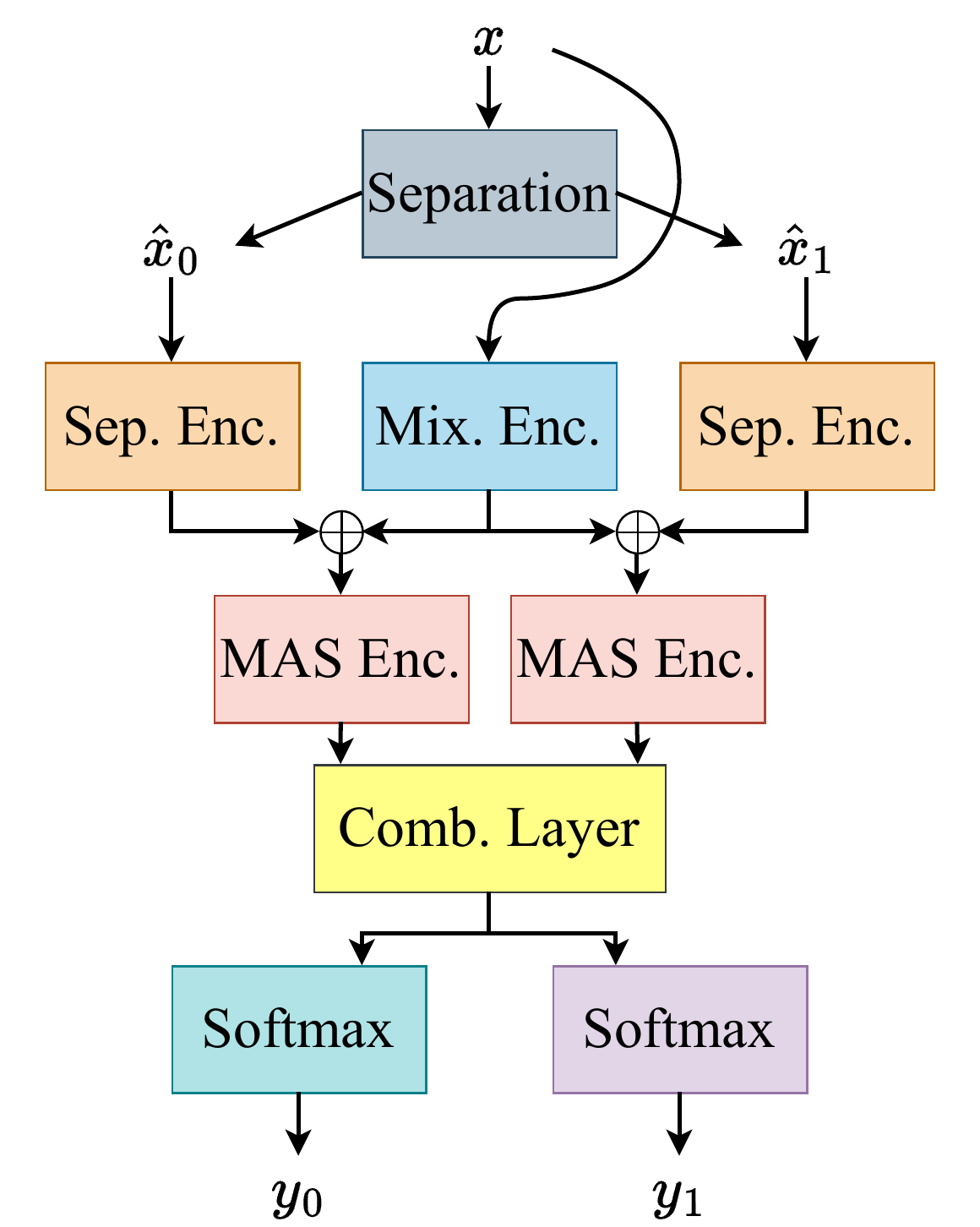}
		\caption{Structure with combine layer in addition to \acrshort{MAS} encoders.}
		\label{fig:combine}
	\end{subfigure}
	\caption{Full structure of multi-speaker model variants. Same coloring indicates shared parameters.}
\end{figure*}

\section{Related work}

There has been significant research that focuses on a speech separation frontend which is used together with an \gls{ASR} backend in a modular way for multi-speaker \gls{ASR} \cite{isik2016modular, wang2022tfGridNet}.
Several studies have shown that a joint fine-tuning of the speech separation- and \gls{ASR}-module improves the performance of such modular systems \cite{neumann2020joint, settle2018modular, qian2018single, zhang2022joint, neumann2020multitalker}.

End-to-end multi-speaker \gls{ASR} systems that do not require explicit speech separation have also been investigated \cite{seki2018purely, sklyar2021msrnnt, sklyar2022mtrnnt}.
These approaches usually either employ a mixture encoder followed by two speaker-dependent encoders \cite{seki2018purely, sklyar2021msrnnt, sklyar2022mtrnnt, chang2019e2e} or they include an explicit mask-based speech separation module, the output of which is applied to the mixture encoder \cite{tripathi2020e2e, lu2021streaming, lu2021multi}.
Such end-to-end approaches train the entire network using an \gls{ASR} criterion and utilize assignment strategies such as \gls{PIT} \cite{yu2017pit}, \gls{SOT} \cite{kanda2020sot}, \gls{DAT} \cite{sklyar2021msrnnt} or \gls{HEAT} \cite{lu2021streaming} to solve the permutation problem.\par
In the field of diarization on audio with overlapping speech, the target speaker-voice activity detection (TS-VAD) model \cite{medennikov2020tsvad} has shown an impressive performance in the CHiME-6 challenge \cite{watanabe2020chime}.
For each speaker embedding vector at the input, it predicts the temporal activity of the respective speaker.
The key for its success is a final combination layer that encodes all speaker information simultaneously while the preceding layers only see one speaker embedding vector each.
In this work, we adapt this idea for multi-speaker \gls{ASR}.

To the best of our knowledge, this is the first work that utilizes an explicit pre-trained speech separation network in a modular way while also leveraging the mixed audio directly through a mixture encoder inside the \gls{ASR} module and allows the flow of cross-speaker context information through a combination layer.

\section{Methods}

\subsection{Speech Separator}
The input to our system is a mixture audio $x$ which consists of multiple overlapping noisy and reverberant single speaker signals.
The speech separation module produces estimates of the single speaker audios $\hat{x}_s$ whereby we only consider the two-speaker scenario, i.e.\@{} $s\in\{0, 1\}$.

To achieve this, we use a mask-based \gls{NN} source separator in the \gls{STFT} domain.
This means, a \gls{NN} is trained to predict one mask for each speaker and then the masks are multiplied with the observation to obtain estimates for each speaker's signal.
After an inverse \gls{STFT}, we use a \gls{SDR}-based loss \cite{heitkaemper2020demystifying} in \gls{PIT} \cite{kolbaek2017uPITBLSTM} to train the \gls{NN}.
To be more specific, a soft upper bound \cite{wisdom2020mixtureOfMixture} for the \gls{SDR} is used, to reduce the contribution of easy examples to the gradient-based optimization.

In recent years, new architectures have been proposed showing superior performance on reverberation-free data \cite[Table 1]{subakan2021SepFormer}.
However, for reverberated data a simple architecture like multiple \gls{BLSTM} layers \cite{kolbaek2017uPITBLSTM} combined with a time domain loss \cite{heitkaemper2020demystifying} still exhibits similar performance, as it is shown in \cite{cord2022monauralSep}.





\subsection{Acoustic Model}

This section discusses the different composition possibilities for a network leveraging both the individual speaker audios $\hat{x}_0$ and $\hat{x}_1$ produced by a speech separation module, as well as incorporating the mixed audio $x$ directly to produce output labels $y_0$ and $y_1$ for the two speakers.

\Cref{fig:modular} shows a purely modular \gls{AM} structure that recognizes the separated audio relying only on the separator output.
This baseline approach is modeled by the following equations:
\begin{align}
	z^\textit{Sep}_s                   & = \textit{SepEnc}(\hat{x}_s)       \\
	p(y_s \mid \hat{x}_s) & = \text{softmax}(W \cdot z^\textit{Sep}_s + b). 
\end{align}
Here, $s \in \lbrace 0, 1 \rbrace$ represents the speaker index, $\textit{SepEnc}$ refers to the separation encoder and $W$ and $b$ are weight and bias of a single linear layer that projects to the dimension of the label space.

If we choose to incorporate $x$ directly, various different possibilities of encoder compositions arise.
\Cref{fig:nocombine} shows a structure that includes the separation encoders that operate on $\hat{x}_0$ and $\hat{x}_1$, as well as a mixture encoder that receives $x$ as its input.
The outputs of the separation encoders and the mixture encoder are added element-wise before being fed into a \gls{MAS} encoder $\textit{MASEnc}$ to produce a common encoding that captures $\hat{x}_s$ and $x$.
The equations for this approach are as follows:
\begin{align}
	z^\textit{MAS}_s        & = \textit{MASEnc}\left(\textit{SepEnc}(\hat{x}_s) + \textit{MixEnc}(x) \right) \\
	p(y_s \mid \hat{x}_s, x) & = \text{softmax}(W \cdot z^\textit{MAS}_s + b).                               
\end{align}
To investigate the individual contributions, we also explore composition options where certain encoders are disabled, i.e.\@{} replaced by an identity function (see \Cref{table:results_mix}).
To handle dimensional mismatches, we use concatenation instead of element-wise addition for $\textit{SepEnc}(\hat{x}_s)$ and $\textit{MixEnc}(x)$ if either of these encoders is replaced by identity.

We investigate one more configuration option, which is the addition of a combination layer (Figures~\ref{fig:nomas_combine} and \ref{fig:combine}) that processes encodings of all three signals $\hat{x}_0$, $\hat{x}_1$ and $x$ jointly.
This option is inspired by \cite{medennikov2020tsvad} where such a combination layer is employed for the TS-VAD model in diarization of overlapping speech.
The motivation for this combination layer is to enable an \enquote{information exchange} between the two halves of the model and constitute a form of soft context between the speakers.
Further variants that exploit cross-speaker context are a topic for future research.
In one variant we omit the \gls{MAS} encoder and directly concatenate ($\parallel$) the outputs of the three preceding encoders (\Cref{fig:nomas_combine}) in order to form the input of the combination layer:
\begin{align}
	  & z^\textit{Comb}                    = Comb\left(z^\textit{Sep}_0 \Vert \textit{MixEnc}(x) \Vert z^\textit{Sep}_1\right) \\
	  & p(y_s \mid \hat{x}_0, x, \hat{x}_1) = \text{softmax}(W_s \cdot z^\textit{Comb} + b_s).                                                     
\end{align}
Here, $Comb$ represents the combination layer and $W_s$, $b_s$ are the parameters of a speaker-specific linear layer.
Two separate sets of parameters for the linear layers are required since they share the same input $z^\textit{Comb}$ which jointly encodes information for both speakers.

Our final variant (\Cref{fig:combine}) is similar to the previous one but we keep the \gls{MAS} encoders and concatenate their outputs to form the input of the combination layer:
\begin{align}
	z^\textit{Comb} = Comb\left(z^\textit{MAS}_0 \Vert z^\textit{MAS}_1 \right). 
\end{align}


\subsection{Training Strategy}
Our training process is divided into three phases. In the first phase, we pre-train the speech separation module independently using an \gls{SDR}-loss (see \cite{cord2022monauralSep}).
During the second phase, we freeze the separator and incorporate our acoustic model, which is optimized using a frame-wise \gls{CE} loss.
To obtain the targets needed for this loss function, we align the clean, un-mixed utterances with a \gls{GMM}.
To ensure that the alignments are compatible with the mixed utterances, we pad them with silence-labels analogous to the zero-padding for the waveforms before mixing.
In the third phase, we unfreeze the separator and jointly optimize it together with the \gls{AM} using the same frame-wise \gls{CE} loss.
Similar to previous works such as \cite{neumann2020joint, qian2018single, zhang2022joint, neumann2020multitalker}, this joint optimization approach can be expected to result in improved performance.

\section{Experimental Setup}
\subsection{Dataset}
For the experiments, we used the SMS-WSJ \cite{drude2019smswsj} dataset.
It uses the WSJ0 and WSJ1 utterances \cite{paul1992wsj}, down-sampled to \SI{8}{\kilo\hertz}, and reverberates them with simulated \glspl{RIR} \cite{allen1979image} with a sound decay time $T_{60}$ in the range of \SI{0.2}{\second} to \SI{0.5}{\second}.
To create a mixture, two reverberated WSJ utterances are added, where the shorter utterance is padded to the same length as the longer utterance.
Further, Gaussian noise with an average \gls{SNR} of \SI{25}{\decibel} is added to the mixture.
While this dataset contains multiple microphones, we only use the first microphone in this work.

\subsection{Speech Separation Model}
For the separation model, we use a re-implementation of the \gls{PIT}-\gls{BLSTM} from \cite{cord2022monauralSep}, with 3 BLSTM layers followed by 2 feed-forward layers.
With the baseline \gls{ASR} model from SMS-WSJ \cite{drude2019smswsj}, this separator achieves a \gls{WER} of \SI{32.83}{\percent} on eval'92, which can be compared to the \SI{35.70}{\percent} from \cite{cord2022monauralSep}.
For the pre-training, we use dynamic mixing \cite{zeghidour2021wavesplit}.
This means, we sample new speaker combinations on demand and reverberate them with the precomputed \glspl{RIR}.


\subsection{Acoustic Model Training}
To train a full \gls{ASR} model, we first employ a training phase where the parameters of the speech separation module are frozen and only the \gls{AM} is trained.
The output of the separator is in the waveform domain and we extract 40-dimensional Gammatone features for both the overlapped and separated audios at the beginning of the separation and mixture encoders.
We use a hybrid \gls{HMM} based acoustic model.
The model utilizes \gls{BLSTM} encoders, which have varying layer counts for each encoder variant.
The model's output alphabet is comprised of $9001$ CART labels.
The \gls{BLSTM} layer width is $400$ for all encoder types and $800$ for the combination layer.
During training, we add an auxiliary softmax output directly on the separation encoders, which optimizes a \gls{CE} loss with the same targets as the final output layers and a scale of 0.3.
This is done to aid in better convergence.
To solve the problem of assigning the correct target sequences to the two softmax outputs, we use the \gls{STFT} magnitude of the reference reverberant signals and of the separated signals and choose the permutation that minimizes the \gls{MSE} between them.
We train the model for $20$ epochs using a Newbob \gls{LR} schedule with an initial \gls{LR} of $4e^{-4}$ and employ $L2$ weight decay of factor $1e^{-2}$, $10\%$ dropout and $0.1$ gradient noise as regularization techniques.
During this phase, we also utilize SpecAugment \cite{park2019specaug}, a data augmentation technique that randomly masks time and frequency blocks of the input features.
The training is performed using the RETURNN toolkit \cite{zeyer2018returnn}.

\subsection{Joint Training}
After individually pre-training the speech separation module and acoustic model, we optimize them jointly in a third training phase.
We unfreeze the parameters of the speech separator and continue training with the frame-wise \gls{CE} loss for another 20 epochs, using a Newbob \gls{LR} schedule and an initial \gls{LR} of $3e^{-5}$.
We found that SpecAugment was not helpful in this training phase and therefore disable it.

\subsection{Recognition}
Recognition is carried out using the RASR toolkit \cite{rybach2011RASR}
and utilized a $3$-gram \gls{LM}.
To recognize the overlapped speech, one speaker was recognized at a time, i.e.\@{} two recognition runs are performed with only one of the softmax outputs being enabled in each run.
After recognition, the hypothesized transcriptions are assigned and scored against the reference transcriptions in the permutation that minimized the \gls{WER}.

\section{Results}

\input{tables/results_mix}

\Cref{table:results_mix} shows the results of our experiments with different encoder sizes and compositions.
Our modular baseline system with $6$-layer separation encoders achieves a \gls{WER} of $20.4\%$ on dev'93 and $14.6\%$ on eval'92.
We observe that purely increasing the size of the modular system does not yield any improvement but in fact leads to a small degradation in performance, with a \gls{WER} of $20.8\%$ on dev'93 and $14.9\%$ on eval'92.
This can be attributed to the increased difficulty in training the larger model and the higher risk of overfitting.

Removing the separation encoders clearly underperforms, with a \gls{WER} of $21.7\%$ on dev'93 and $15.1\%$ on eval'92.
One possible explanation for this is that the cleaner audio produced by the speech separator is not leveraged enough in this model structure.
However, in all other cases, the inclusion of mixed audio through a mixture encoder and/or \gls{MAS} encoder led to an improvement over the baseline.
Without a combination layer, including both mixture and MAS encoder does not lead to a significant improvement over having only one of them active, so the increased model size in this variant is not justified by the performance and only including one of them is sufficient.
Furthermore, adding a combination layer requires a large increase in the number of model parameters due to the concatenated inputs, but it results in the best performance, reducing the \gls{WER} to $19.1\%$ on dev'93 and $13.6\%$ on eval'92.
This represents a relative improvement of about $7\%$ over the baseline.

Overall, our experiments demonstrate that incorporating mixed audio through a mixture or \gls{MAS} encoder, along with a combination layer to enable the sharing of cross-speaker context information, can significantly improve the performance of our \gls{ASR} system.

\section{Discussion}

As demonstrated by our results in \Cref{table:results_mix}, our novel approach to acoustic modeling in multi-speaker \gls{ASR} results in clear improvements over the baseline method, which cannot be replicated by simply making the baseline network larger.
Because of the parameter sharing and the way that mixture encoder and separation encoders are combined, the structure with mixture encoder and/or MAS encoder (\Cref{fig:nocombine}) can easily be applied to an arbitrary amount of speakers.
However, the extension of the combination layer to an arbitrary speaker count is not trivial and requires further investigation.
Our performance increase using the combination layer suggests that more exploration of cross-speaker context may be a valuable topic for future research.

Our proposed encoder configuration could also be applied to other \gls{AM} architectures such as the \gls{RNN-T}, which opens up further research opportunities.

\input{tables/results_literature_wsj}

\input{tables/results_literature_smswsj}
Finally, our methods are compared to results reported in the literature.
\Cref{table:results_literature_wsj} shows a comparison on the clean WSJ corpus.
Here, we train and evaluate an acoustic model with the same configuration as the \gls{AM} used as our baseline with one encoder followed by a softmax output (see \Cref{fig:modular}).
This model shows similar performance to the \acrlong{SOTA} presented in \cite{hadian2018LFMMI}.

\Cref{table:results_literature_smswsj} shows results on the SMS-WSJ corpus.
For these results, it is worth noting that the \gls{ASR} performance of modular multi-speaker systems is highly dependent on the quality of the speech separation module, making direct comparisons of the \gls{AM} with literature results difficult.
Nevertheless, it is clear that the proposed model with mixture encoding outperforms other recent approaches \cite{wang2021multi, cord2022monauralSep} and even multi-channel techniques like \cite{zhang2022joint} despite using only a single channel.
Only the TF-GridNet separator presented in \cite{wang2022tfGridNet} outperforms all other reported approaches by a large margin even using the baseline Kaldi \gls{ASR} backend.
This supports the effectiveness of their separator.
In contrast, we use a rather standard speech separator and focus on improving acoustic modeling for \gls{ASR}.
Future work could include a combination of the separator from \cite{wang2022tfGridNet} with our \gls{ASR} backend.

\section{Conclusions}

In this research, we propose a middle-ground approach for multi-speaker \gls{ASR} that combines the benefits of explicit speech separation with the direct incorporation of mixture speech information into a hybrid \gls{HMM}-based \gls{ASR} module.
Our approach also includes a layer that combines encoder information of individual speakers to exchange cross-speaker context information.
The system is optimized through separate and joint training stages.

Our experiments show that the proposed approach outperforms the conventional modular baseline method by around $7\%$ relative improvement in \gls{WER}.
Our proposed method of incorporating mixture audio can easily be applied to a higher speaker count and the general encoder structure can also be used for other \gls{AM} architectures, such as \gls{RNN-T}, which opens up opportunities for future research.

The inclusion of a combination layer which merges the encoder data of individual speakers to share cross-speaker context information, drawing inspiration from work in speaker diarization, also clearly improves the \gls{ASR} accuracy.
This finding indicates that more research on cross-speaker context may prove valuable in the future.
Overall, our approach is a promising solution for multi-speaker \gls{ASR} that utilizes explicit speech separation, creates fertile opportunities for further extensions and investigations, and has the potential to enhance performance in real-world applications.

\section{Acknowledgements}

\ifinterspeechfinal
This research was partially supported by the Deutsche Forschungsgemeinschaft (DFG) under project no. 448568305.  
\else
Acknowledgements hidden for review.
\fi

\bibliographystyle{IEEEtran}
\bibliography{mybib}

\end{document}

%% file: acronyms.tex
\newacronym{AM}{AM}{acoustic model}
\newacronym{AMI}{AMI}{Augmented Multi-party Interaction}
\newacronym{ARQ}{ARQ}{automatic repeat request}
\newacronym{ASR}{ASR}{automatic speech recognition}
\newacronym[longplural={bi-directional long-short term memories}]{BLSTM}{BLSTM}{bi-directional long-short term memory}
\newacronym{BSS}{BSS}{blind speech separation}
\newacronym{CART}{CART}{classification and regression tree}
\newacronym{CE}{CE}{cross entropy}
\newacronym{CER}{CER}{character error rate}
\newacronym{CDp}{CDp}{context dependent phoneme}
\newacronym{CNN}{CNN}{convolutional neural network}
\newacronym{CPC}{CPC}{contrastive predictive coding}
\newacronym{CTC}{CTC}{connectionist temporal classification}
\newacronym{DAT}{DAT}{deterministic assignment training}
\newacronym{DCT}{DCT}{discrete cosine transform}
\newacronym{DL}{DL}{deep learning}
\newacronym{DNN}{DNN}{deep neural network}
\newacronym{DNN-HMM}{DNN-HMM}{deep neural network hidden Markov model}
\newacronym{ELBO}{ELBO}{evidence lower bound}
\newacronym{EM}{EM}{expectation maximization}
\newacronym{FE}{FE}{feature extractor}
\newacronym{FIR}{FIR}{finite impulse response}
\newacronym{FFNN}{FFNN}{feed-forward neural network}
\newacronym{fCE}{fCE}{frame-wise cross-entropy}
\newacronym{G2P}{G2P}{grapheme-to-phoneme conversion}
\newacronym{GAN}{GAN}{generative adversarial network}
\newacronym{GMM}{GMM}{Gaussian mixture model}
\newacronym{GMM-HMM}{GMM-HMM}{Gaussian mixture model hidden Markov model}
\newacronym{GPU}{GPU}{graphics processing unit}
\newacronym{HEAT}{HEAT}{heuristic error assignment training}
\newacronym{HMM}{HMM}{hidden Markov model}
\newacronym{IHM}{IHM}{individual headset microphones}
\newacronym{IIR}{IIR}{infinite impulse response}
\newacronym{LAS}{LAS}{listen-attend-spell}
\newacronym{LC-BLSTM}{LC-BLSTM}{latency-controlled bidirectional long-short term memory}
\newacronym{LDA}{LDA}{linear discriminant analysis}
\newacronym{LM}{LM}{language model}
\newacronym[longplural={long-short term memories}]{LSTM}{LSTM}{long-short term memory}
\newacronym{LR}{LR}{learning rate}
\newacronym{MAS}{MAS}{mixture-aware speaker}
\newacronym{MDM}{MDM}{multiple distant microphones}
\newacronym{MFCC}{MFCC}{Mel-frequency cepstral coefficients}
\newacronym{MHSA}{MHSA}{multi-head self-attention}
\newacronym{MRES}{MRES}{multi-resolutional}
\newacronym{MSE}{MSE}{mean squared error}
\newacronym{MT}{MT}{machine translation}
\newacronym{NLP}{NLP}{natural language processing}
\newacronym{NN}{NN}{neural network}
\newacronym{OOV}{OOV}{out-of-vocabulary}
\newacronym{PC2}{$\text{PC}^{\text{2}}$}{Paderborn Center for Parallel Computing}
\newacronym{PIT}{PIT}{permutation invariant training}
\newacronym{RIR}{RIR}{room impulse response}
\newacronym{RNA}{RNA}{recurrent neural aligner}
\newacronym{RNN}{RNN}{recurrent neural network}
\newacronym{RNN-T}{RNN-T}{recurrent neural network transducer}
\newacronym{RSAN}{RSAN}{recurrent selective attention network}
\newacronym{SAT}{SAT}{speaker adaptive training}
\newacronym{SDM}{SDM}{single distant microphone}
\newacronym{SDR}{SDR}{signal-to-distortion ratio}
\newacronym{SC}{SC}{supervised convolutional}
\newacronym{SNR}{SNR}{signal to noise ratio}
\newacronym{sMBR}{sMBR}{state-level minimum Bayes risk}
\newacronym{SOT}{SOT}{serialized output training}
\newacronym{SOTA}{SOTA}{state of the art}
\newacronym{STFT}{STFT}{short time Fourier transform}
\newacronym{TDNN}{TDNN}{time delay neural network}
\newacronym[longplural={time-frequencies}]{tf}{tf}{time-frequency}
\newacronym{VAD}{VAD}{voice activity detection}
\newacronym{VTLN}{VTLN}{vocal tract length normalization}
\newacronym{cpWER}{cpWER}{concatenated minimum-permutation word error rate}
\newacronym{WER}{WER}{word error rate}
\newacronym{WERR}{WERR}{word error rate reduction}
\newacronym{WP}{WP}{work package}
\newacronym{WPE}{WPE}{weighted prediction error}
\newacronym{WSJ}{WSJ}{Wall Street Journal}

%% file: tables/results_mix.tex
\begin{table}[htbp]
																							
	\centering
	\caption{Performance of different \glspl{AM} structures. Shown are the number of \gls{BLSTM} layers of individual components, total number of model parameters (including speech separator) and \glspl{WER}.}
	\label{table:results_mix}
	\begin{tabular}{|c|c|c|c|c|c|c|}
		\hline
		\multicolumn{4}{|c|}{Num encoder layers} & \# Par. & \multicolumn{2}{c|}{WER [\%]}                                                                           \\\cline{1-4}\cline{6-7}
		Sep                & Mix                & MAS                & Comb               & [Mil.] & dev'93        & eval'92       \\\hline\hline
		6                  & \multirow{2}{*}{-} & \multirow{2}{*}{-} & \multirow{5}{*}{-} & \ 51   & 20.4          & 14.6          \\\cline{1-1}\cline{5-7}
		8                  &                    &                    &                    & \ 80   & 20.8          & 14.9          \\\cline{1-3}\cline{5-7}
		-                  & 6                  & \multirow{2}{*}{4} &                    & \ 67   & 21.7          & 15.1          \\\cline{1-2}\cline{5-7}
		\multirow{5}{*}{6} & -                  &                    &                    & \ 74   & 19.6          & 14.1          \\\cline{2-3}\cline{5-7}
		                   & \multirow{4}{*}{4} & \multirow{2}{*}{-} &                    & \ 64   & 19.9          & 14.1          \\\cline{4-7}
		                   &                    &                    & 1                  & 114    & 19.5          & 13.8          \\\cline{3-7}
		                   &                    & 2                  & -                  & \ 80   & 19.6          & 14.0          \\\cline{3-7}
		                   &                    & 1                  & 1                  & 120    & \textbf{19.1} & \textbf{13.6} \\\hline
	\end{tabular}
																							
\end{table}

%% file: tables/results_literature_wsj.tex
\begin{table}[htbp]
					
	\centering
	\caption{Comparison of results reported in the literature for the clean WSJ corpus.}
	\label{table:results_literature_wsj}
	\begin{tabular}{|c|c|c|S|S|}
		\hline
		\multirow{2}{*}{Model}   & \multicolumn{2}{c|}{WER [\%]}           \\\cline{2-3}
		                              & {dev'93} & {eval'92} \\\hline\hline
		LF-MMI \cite{hadian2018LFMMI} & {-}      & 2.9       \\\hline
		ours                          & 5.4      & 3.0       \\\hline
	\end{tabular}
					
\end{table}

%% file: tables/results_literature_smswsj.tex
\begin{table}[htbp]
						
	\centering
	\caption{Comparison of results reported in the literature for SMS-WSJ corpus. \#Ch denotes the number of microphone channels used.}
	\label{table:results_literature_smswsj}
	\begin{tabular}{|c|c|c|S|S|}
		\hline
		\multirow{2}{*}{Model}   & \multirow{2}{*}{\#Ch} & \multicolumn{2}{c|}{WER [\%]}           \\\cline{3-4}
		                                        &                    & {dev'93} & {eval'92} \\\hline\hline
		Joint Sep. \& ASR \cite{zhang2022joint} & 6                  & {-}      & 15.4      \\\hline
		2-spkr ASR \cite{zhang2022joint}        & \multirow{5}{*}{1} & {-}      & 35.0      \\\cline{1-1}\cline{3-4}
		U-Net TCN \cite{wang2021multi}          &                    & {-}      & 28.7      \\\cline{1-1}\cline{3-4}
		SepFormer \cite{cord2022monauralSep}    &                    & {-}      & 26.5      \\\cline{1-1}\cline{3-4}
		TF-GridNet \cite{wang2022tfGridNet}     &                    & {-}      & 7.9       \\\cline{1-1}\cline{3-4}
		ours                                    &                    & 19.1     & 13.6      \\
		\hline
	\end{tabular}
						
\end{table}